\documentclass[conference]{IEEEtran}
\IEEEoverridecommandlockouts

\usepackage{cite}
\usepackage{amsmath,amssymb,amsfonts}
\usepackage{algorithmic}
\usepackage{graphicx}
\usepackage{textcomp}
\usepackage{xcolor}
\usepackage{booktabs}
\usepackage{url}
\usepackage{hyperref}
\def\BibTeX{{\rm B\kern-.05em{\sc i\kern-.025em b}\kern-.08em
    T\kern-.1667em\lower.7ex\hbox{E}\kern-.125emX}}

\begin{document}
 
\title{MASH-Bench: Diagnosing Cross-Source Failure in Mass-Shooting Risk Classification.}
 
\author{
% \IEEEauthorblockN{Anonymous Authors}
\IEEEauthorblockN{Neha Sharma}
\IEEEauthorblockA{\textit{Department of Computer Science} \\
\textit{Virginia Commonwealth University}\\
Richmond, Virginia, United States\\
sharman20@vcu.edu}
\and
\IEEEauthorblockN{Ritesh Sharma}
\IEEEauthorblockA{\textit{Department of Electrical and Computer Engineering} \\
\textit{Virginia Commonwealth University}\\
Richmond, Virginia, United States\\
sharmar33@vcu.edu}
}
 
\maketitle
 
\begin{abstract}
Public mass-shooting databases differ substantially in coverage, feature availability, and reporting practices, creating challenges for machine-learning models that must generalize across data sources. We introduce \textbf{MASH-Bench}, a harmonized benchmark of 6,968 incidents from four U.S. databases---Kaggle, Mother Jones, Stanford MSA, and the Gun Violence Archive (GVA)---and evaluate cross-source risk classification using leave-one-dataset-out (LODO) evaluation. Random Forest, XGBoost, and LightGBM achieve VeryHigh-risk recall of 0.68--0.89 on the curated sources but generalize poorly to GVA, where mean recall drops to 0.20 and precision to 0.0004. To investigate the source of this degradation, we conduct a controlled feature-masking ablation that removes the five features unavailable in GVA from the curated sources. The resulting recall collapse to zero provides evidence that feature completeness is a major contributor to the observed cross-source failure. We further evaluate three domain-adaptation approaches: DANN, CORAL, and importance weighting. DANN improves VeryHigh-risk recall on GVA by 0.282 (95\% CI [0.11, 0.47], $p = 0.003$), although precision remains low, whereas CORAL and importance weighting yield zero recall. Oracle prior-shift recalibration likewise fails to recover VeryHigh-risk predictions, indicating that label-side correction alone is insufficient under the observed feature deficiencies. A per-group audit further identifies substantial disparities associated with media-attributed mental-health labels. Overall, these results indicate that, in MASH-Bench, cross-source generalization is constrained more by feature completeness and label prevalence than by classifier choice. The benchmark provides a controlled setting for diagnosing these effects in cross-source risk classification.
\end{abstract}

 
\section{Introduction}

Machine learning studies based on public data increasingly rely on multiple databases that differ substantially in coverage, inclusion criteria, and reporting practices. Mass-shooting research provides a concrete example of this cross-source heterogeneity. Journalistic sources typically apply editorial criteria when selecting incidents, academic databases may impose thresholds based on fatalities or other incident characteristics, and open-source repositories aggregate records from law-enforcement, media, and other public sources. These differences can support valid within-source analyses while creating uncertainty about whether models trained on one source generalize to another. Despite the growing use of machine learning for risk classification, cross-source generalization has received limited systematic evaluation.

Prior work by Sharma et al.~\cite{sharma2026harmonizing} investigated cross-source mass-shooting risk classification across three curated sources---Kaggle, Mother Jones, and Stanford MSA---covering 806 incidents. That study established a common representation and found no statistically significant difference in cross-source performance among the evaluated feature configurations and stratification strategies. However, the strongest configuration was partially influenced by a distributional artifact in the Mother Jones data. Two limitations motivated the present study. First, the relatively small pooled sample limited the ability to distinguish systematic differences between sources from sampling variability. Second, the statistical analysis did not account for dependence among the 27 evaluated configurations, limiting the reliability of inference for the feature-set comparison.

We address these limitations by introducing MASH-Bench, an expanded benchmark that adds the Gun Violence Archive (GVA)~\cite{gva} to the three curated sources, yielding 6,968 incidents and substantially expanding coverage of the post-2013 period. The expanded benchmark reveals a pronounced cross-source generalization gap. Models trained on the three curated sources achieve VeryHigh-risk recall between 0.68 and 0.89 under leave-one-dataset-out (LODO) evaluation, but performance degrades substantially when transferring to GVA, where mean VeryHigh-risk recall is 0.20 and precision is 0.0004. Rather than treating poor transfer as simply a difference between classifiers, we use controlled experiments to diagnose the underlying source of failure. A feature-masking ablation (Section~\ref{sec:masking}) provides evidence for the feature-completeness mechanism: masking the five features GVA lacks on curated sources collapses their recall to zero, reproducing the qualitative failure pattern observed on GVA. Among the evaluated domain-adaptation methods, only domain-adversarial neural network (DANN) training produces a statistically significant improvement in GVA recall, whereas covariance alignment (CORAL) and importance weighting do not. Oracle prior-shift recalibration likewise fails to recover the missing signal, indicating that label-side correction alone is insufficient when the target feature representation is incomplete.

The main contributions of this work are as follows:

\begin{itemize}
\item \textbf{MASH-Bench}: We introduce a harmonized benchmark spanning four public U.S.\ mass-shooting databases, comprising 6,968 incidents and a common 12-variable schema for cross-source risk classification.

\item \textbf{Cross-source generalization analysis and mechanism ablation}: We evaluate Random Forest, XGBoost, and LightGBM under LODO evaluation and show that models achieving strong recall on the curated sources (0.68--0.89) generalize poorly to GVA. A controlled feature-masking ablation provides evidence that feature completeness is a primary contributor: masking the five features GVA lacks collapses curated-source recall to zero.

\item \textbf{Domain adaptation analysis}: We compare DANN, CORAL, and importance weighting under a common cross-source protocol and characterize their benefits and failure modes under feature heterogeneity.

\item \textbf{Inference under configuration dependence}: We use clustered bootstrap confidence intervals and cluster-robust regression to account for dependence among experimental configurations, providing dependence-aware inference for the feature-set and domain-adaptation comparisons.

\item \textbf{Per-group audit}: We examine VeryHigh-risk recall and precision across groups defined by media-attributed mental-health labels, identifying substantial differences in prediction behavior.
\end{itemize}

Overall, MASH-Bench is intended as a diagnostic benchmark rather than a deployable risk-prediction system. It provides a controlled setting for identifying whether cross-source failures arise from model capacity, feature availability, or distributional differences. The observed group-level disparities further underscore the need for careful evaluation before such models are considered for operational use.

\section{Related Work}

\textbf{Predictive modeling of mass shootings.}
Prior machine-learning studies of mass-shooting risk have largely
focused on analyses within individual data sources, including
fatality prediction and descriptive characterization of incidents
~\cite{violence_project}. Related work on machine-learning-based
violence risk assessment has similarly emphasized within-dataset
evaluation~\cite{parmigiani2022}. These studies provide useful
predictive and descriptive analyses but do not directly address whether
models transfer across heterogeneous incident databases.

\textbf{Cross-source generalization in adjacent domains.}
Cross-source evaluation has revealed substantial degradation in model
performance across several domains, including clinical imaging
~\cite{zech2018}, pedestrian detection under domain shift
~\cite{hasan2022pedestrian}, network intrusion detection
~\cite{cantone2024machine}, and tabular benchmarks under distribution
shift~\cite{gardner2023benchmarking}. These studies show that
performance measured within a single source can overestimate performance
when models are transferred to data collected under different
conditions. In network intrusion detection, for example, cross-source
evaluation can reduce classification performance to levels close to
random chance~\cite{cantone2024machine}. MASH-Bench extends this line of
work to heterogeneous public mass-shooting databases by explicitly
evaluating transfer across sources.

\textbf{Definitional heterogeneity in mass-shooting databases.}
Reported mass-shooting incident counts differ substantially across
public databases, in part because the sources use different definitions
and inclusion criteria~\cite{fox_delateur_2014,rocque2018rampage}. These
differences introduce source heterogeneity beyond variation in recorded
features and motivate a common representation when evaluating models
across databases. More broadly, cross-domain data fusion approaches
distinguish stage-based, feature-level, and semantic-meaning-based
fusion~\cite{zheng2015fusion}. MASH-Bench adopts a feature-level
harmonization strategy by mapping source-specific variables into a
common schema.

\textbf{Domain adaptation for tabular data.}
Domain-adaptation methods provide established approaches for mitigating
distributional differences between training and target domains.
CORAL~\cite{sun2016coral} aligns second-order feature statistics,
importance weighting adjusts source examples according to their
relevance to the target distribution~\cite{sugiyama2007importance}, and
DANN~\cite{ganin2015dann} learns representations that reduce domain
discriminability through adversarial training. These approaches address
different forms of source shift and provide complementary baselines for
evaluating whether cross-source performance degradation can be mitigated
in MASH-Bench.

\textbf{Class-imbalance handling.}
Severe class imbalance is common in risk-classification problems and
can substantially affect recall and precision for rare outcomes.
Established approaches include resampling and threshold-selection
methods~\cite{chawla2002smote,youden1950}. In MASH-Bench, class imbalance
is particularly important because the prevalence of the VeryHigh-risk
class differs substantially across sources, making it necessary to
distinguish classifier performance from source-level prevalence effects.

\textbf{Fairness in risk assessment.}
Prior work on algorithmic fairness has examined the use of demographic
attributes and their proxies in predictive models~\cite{barocas_selbst}.
Following this perspective, we conduct a per-group audit by examining
VeryHigh-risk recall and precision across groups defined by
media-attributed mental-health labels. This analysis is used to
identify differences in prediction behavior and performance across
groups.

Although prior work provides established tools for cross-source
evaluation, data harmonization, domain adaptation, and imbalance
handling, these approaches have generally been studied separately.
MASH-Bench combines feature harmonization with controlled
feature-masking, label-shift, and domain-adaptation analyses to diagnose
cross-source failure in heterogeneous mass-shooting databases.

\section{MASH-Bench: Data and Harmonization}
\label{sec:data}

\subsection{Sources}

MASH-Bench harmonizes four public U.S. mass-shooting databases
(Table~\ref{tab:sources}). The three curated sources (Kaggle, Mother
Jones, Stanford MSA) are carried forward from prior work~\cite{sharma2026harmonizing}; GVA~\cite{gva} is the fourth
source added in this study.

\begin{table}[t]
\caption{The four sources in MASH-Bench (total\_victims $\geq 3$; Las
Vegas 2017 excluded from curated sources). SD = standard deviation of
total\_victims.}
\vspace{-0.25cm}
\label{tab:sources}
\centering
\begin{tabular}{lrrrrr}
\toprule
Source & $n$ & Years & Median & Mean & SD \\
\midrule
Kaggle         &  315 & 1966--2017 &  5.0 &  8.56 &  8.8 \\
Mother Jones   &  157 & 1982--2026 & 10.0 & 13.80 & 12.3 \\
Stanford MSA   &  334 & 1966--2016 &  5.0 &  7.53 &  7.8 \\
Gun Violence Archive & 6{,}162 & 2013--2026 &  4.0 &  5.07 &  2.9 \\
\midrule
Total (pooled) & 6{,}968 & 1966--2026 &  4.0 &  5.79 &  4.5 \\
\bottomrule
\end{tabular}
\vspace{-0.25cm}
\end{table}

MASH-Bench expands the previous three-source benchmark in scale, source heterogeneity, and coverage~\cite{han2014big}. It increases the volume to 6,968 incidents, adds a fourth source with a different schema and coverage pattern, and retains sources with different definitional and editorial practices.

Two structural differences between GVA and the curated sources are
central to the subsequent analysis. First, GVA's per-incident casualty
counts are substantially lower on average (mean 5.07 vs.\
7.53--13.80 for the curated sources), reflecting its broader coverage
of lower-severity incidents that are less represented in the curated
sources. Second, GVA systematically lacks participant-level features:
\texttt{incident\_area}, \texttt{gender}, \texttt{race},
\texttt{mental\_health}, and \texttt{age} are recorded as ``Unknown''
for every GVA incident because the public GVA feed does not include
participant fields. These structural differences are examined as potential
contributors to the cross-source performance gap.

\subsection{Common Schema}

We map each source into a shared 12-variable schema: \texttt{source},
\texttt{year}, \texttt{fatalities}, \texttt{injured},
\texttt{total\_victims}, \texttt{incident\_area} (eight normalized
categories), \texttt{open\_close}, \texttt{age}, \texttt{gender},
\texttt{race}, \texttt{mental\_health}, and
\texttt{multiple\_shooters}. Mapping details for the three curated
sources follow~\cite{sharma2026harmonizing}; for GVA, the five
participant-level variables are structurally unavailable and are
therefore recorded as ``Unknown'' for all GVA incidents rather than
imputed. The \texttt{multiple\_shooters} variable is inferred from
GVA's suspect sub-counts (total suspects $>1$).

\subsection{Risk Stratification}
\label{sec:strat}

We evaluate three risk stratification schemes over
\texttt{total\_victims}, assigning Low, Medium, High, and VeryHigh
labels. The first uses fixed rule-based thresholds (10, 20, 40). The
second uses standard deviations ($\mu \pm \sigma$,
$\mu + 2\sigma$). The third uses quartiles. Under LODO,
distribution-derived thresholds (std, quartile) are computed only on
the training fold, preventing label leakage. Throughout this work,
``risk'' refers to severity strata derived from total victim counts,
not prediction of future attack occurrence.

\textbf{Quartile degeneracy on the 4-source pool.} Adding GVA shifts
the pooled quartile boundaries to $Q_1=Q_2=4$ and $Q_3=5$, causing the
``Medium'' class to contain no observations because of the discrete
victim-count distribution. We therefore report only the rule and
standard-deviation strategies throughout the four-source experiments;
quartile results are retained only for direct comparison
with~\cite{sharma2026harmonizing}.

\begin{table}[t]
\centering
\caption{VeryHigh-class base rates per source under the two
stratification strategies. Rule uses fixed thresholds; std uses the
training-fold $\mu+2\sigma$ threshold under LODO.}
\label{tab:baserates}
\begin{tabular}{lcc}
\toprule
Source & Rule VeryHigh rate & Std VeryHigh rate \\
\midrule
Kaggle & $2.2\%$ (7/315) & $13.7\%$ (43/315) \\
Mother Jones & $5.1\%$ (8/157) & $35.7\%$ (56/157) \\
Stanford MSA & $1.8\%$ (6/334) & $11.1\%$ (37/334) \\
GVA & $\mathbf{0.03\%}$ (2/6{,}162) & $\mathbf{0.18\%}$ (11/6{,}162) \\
\bottomrule
\end{tabular}
\end{table}

\subsection{Empirical Distributional Shift}
\label{sec:shift}

To quantify source heterogeneity, we compute three distributional-shift
metrics across the six source pairs (Table~\ref{tab:shift}).
Wasserstein-1 distances are computed per numeric feature (year,
total\_victims, age) and averaged. Total-variation distances are
computed per categorical feature and averaged. Maximum Mean Discrepancy
(MMD$^2$) with an RBF kernel and median-heuristic bandwidth on the
one-hot encoded joint feature matrix provides a single-number
joint-distribution comparison.

\begin{table}[t]
\caption{Distributional shift across the six source pairs.
Wasserstein-1 (larger = greater numeric shift), total-variation
distance (0 = identical, 1 = disjoint), MMD$^2$ (larger = greater
joint-distribution shift). Bolded values: smallest per metric.}
\vspace{-0.25cm}
\label{tab:shift}
\centering
\scriptsize
\setlength{\tabcolsep}{4pt}
\begin{tabular}{llrrr}
\toprule
Source A & Source B & Wasserstein & TV & MMD$^2$ \\
\midrule
GVA          & Kaggle       & 7.77 & 0.69 & 0.138 \\
GVA          & Mother Jones & 9.41 & 0.64 & \textbf{0.135} \\
GVA          & Stanford MSA & 8.18 & 0.48 & \textbf{0.135} \\
Kaggle       & Mother Jones & 5.01 & 0.42 & 0.243 \\
Kaggle       & Stanford MSA & \textbf{1.66} & \textbf{0.26} & 0.169 \\
Mother Jones & Stanford MSA & 4.32 & 0.29 & 0.153 \\
\bottomrule
\end{tabular}
\vspace{-0.25cm}
\end{table}

Kaggle and Stanford MSA exhibit the smallest marginal shift
(Wasserstein 1.66, TV 0.26), consistent with their shared provenance
through the Stanford Geospatial Center. A direct event-level check
quantifies this overlap: $90.5\%$ of Kaggle incidents exactly match a
Stanford MSA row on (year, fatalities, injured), rising to $93.1\%$
within their shared time window (1966--2016). Kaggle--Mother Jones and
Stanford--Mother Jones event overlaps are $27.3\%$ and $16.8\%$,
respectively. These overlaps suggest that LODO transfer among the
three curated sources substantially reflects within-family
generalization. The GVA--curated-source comparison therefore provides
the clearest test of transfer to a substantially different source in
this study.

GVA exhibits the largest measured marginal shifts against all three
curated sources (Wasserstein 7.77--9.41, TV 0.48--0.69), coinciding
with its narrower victim-count distribution and structurally Unknown
participant-level features. Interestingly, the largest joint MMD$^2$
occurs for Kaggle--Mother Jones (0.243), not for a GVA pair. This
suggests that joint distributional distance alone does not explain the
GVA transfer failure and motivates the controlled feature-completeness
analysis in Section~\ref{sec:masking}.

\section{Methodology}
\label{sec:method}
 
\subsection{Classification Pipeline}
 
For each combination of stratification strategy, feature set, and
classifier, we recompute labels using training-fold statistics (where
applicable), apply random oversampling to balance the training classes,
and generate class probabilities on the held-out fold. The held-out
source is never used for label construction, model fitting,
oversampling, or threshold estimation. Under the default decision rule,
predictions follow $\hat{y}_i = \arg\max_k P_{ik}$. We additionally
evaluate a per-class Youden threshold variant (\emph{youden\_val}): a
20\% stratified validation split is held out from the training fold,
class-specific thresholds $t_k$ are selected by maximizing
$\text{TPR}_k(t) - \text{FPR}_k(t)$ on that split, and predictions
follow $\hat{y}_i = \arg\max_k \{P_{ik} - t_k\}$. All reported results
use the default and \emph{youden\_val} rules only.
 
\subsection{Experimental Setup}
\label{sec:setup}
 
\textbf{Feature sets.} We evaluate two feature configurations. The
contextual set (ctx) contains incident and demographic context
available across sources (incident\_area, open\_close, gender). The
full set extends this representation with age, mental\_health, race,
and multiple\_shooters. Crucially, neither feature set includes the
numeric outcome variables (fatalities, injured, or total\_victims),
from which risk labels are derived; including any of these would
trivialize the classification task. Year is also excluded from both
feature sets to avoid time-source confounding under LODO evaluation.
 
\textbf{Classifiers.} We evaluate three classifier families:
Random Forest (200 trees, max depth 10, balanced class weights);
XGBoost (300 trees, max depth 6, learning rate 0.1); LightGBM
(matching XGBoost hyperparameters). Where prior-work comparisons are
relevant we reference the depth-3 decision tree, multinomial logistic
regression, and Gaussian na\"ive Bayes of~\cite{sharma2026harmonizing}.
 
\textbf{Domain adaptation.} We evaluate three DA strategies.
\emph{CORAL}~\cite{sun2016coral} applies a whitening re-colouring
transform to align source and target second-order statistics before
classifier training. Because the representation contains many one-hot
categorical variables and GVA assigns ``Unknown'' to all
participant-level fields, CORAL is evaluated as a covariance-based
alignment baseline under substantial categorical sparsity.
\emph{Importance weighting (IW)} estimates target-vs-source density
ratios using a probabilistic domain classifier (Random Forest, 200
trees, depth 8) with per-sample weights clipped to $[0.1, 10]$ and
mean-normalized to preserve effective sample size. \emph{DANN}~\cite{ganin2015dann} uses a
two-layer MLP (hidden width 64) with a gradient-reversal domain head,
jointly trained to maximize label prediction and minimize
source-vs-target domain distinguishability with a Ganin-style adaptive
$\lambda$ schedule ($\lambda = 2/(1+e^{-10p}) - 1$ scaled by
$\lambda_{\max}$, where $p$ is training progress) and 100 max epochs.
The domain classifier is $K$-way where $K = |\text{source domains}| + 1$
(one class index reserved for the target), rather than the binary
formulation of the original DANN; this is the multi-source variant
of~\cite{zhao2018multisource} adapted to the LODO protocol. Using
per-source domain labels encourages the feature extractor to remove
both source-vs.-target and inter-source domain structure, which is
appropriate for the four-source setting in which the curated sources
also differ systematically from one another. Source-domain labels
are re-encoded per LODO fold. Features enter the MLP in the same
one-hot-plus-numeric representation used by the baseline classifiers.
CORAL and IW use XGBoost as the base learner; DANN's MLP replaces this.
All three DA methods are run on the same LODO folds as the baseline
XGBoost pipeline.
 
\textbf{Evaluation protocol.} Within-source evaluation uses stratified
5-fold cross-validation on the pooled 6{,}968-row dataset as a reference
for pooled-data performance; all cross-source claims rely on LODO.
Cross-source evaluation uses leave-one-dataset-out (LODO): models are
trained on three sources and evaluated on the fourth, repeated over five
random seeds (42--46). For categorical variables, the encoder's column
set is the union of category levels appearing in the combined training
and held-out rows, so that column indexing is consistent and no
unseen-category failures occur; this uses only the set of category
levels that can occur, not target observations, label proportions,
feature frequencies, or other statistics from the held-out source.
 
\textbf{Seed-randomness scope.} Under LODO, the held-out source is
fixed and the test partition therefore does not vary across seeds.
Seeds 42--46 vary only (a) the random-oversampling bootstrap on the
training fold, (b) the 20\% stratified validation split used for
youden\_val threshold selection, and (c) tie-breaking or stochastic
optimization within each classifier.
 
\subsection{Inference Protocol}
\label{sec:inference}
 
To evaluate paired comparisons across configurations (ctx-vs-full and
DA-vs-baseline), we use two configuration-dependence-aware inference
methods that account for the dependence induced by shared experimental
configurations. For each paired comparison, $\Delta$ denotes the
difference in VeryHigh recall between configurations evaluated on the
same test source and seed:
$\Delta = \mathrm{Recall}_{full} - \mathrm{Recall}_{ctx}$ for the
feature-set comparison and
$\Delta = \mathrm{Recall}_{DA} - \mathrm{Recall}_{baseline}$ for the
domain-adaptation comparison.
 
\emph{Clustered bootstrap.} We resample the set of paired-config
$\Delta$ values with replacement 10{,}000 times, computing the mean
$\Delta$ in each resample. The 2.5\%--97.5\% quantile range gives a
95\% confidence interval for the mean effect.
 
\emph{Cluster-robust regression.} We fit $\Delta_i = \beta_0 +
\epsilon_i$ via OLS with cluster-robust standard errors, where clusters
are configuration triples (classifier, strategy, threshold\_mode) for
the feature-set comparison and (strategy, features) for the
DA-vs-baseline comparison. These clusters group paired observations
sharing the same experimental configuration and therefore account for
within-configuration dependence across test sources and seeds.
 
Both methods account for the configuration dependence that~\cite{sharma2026harmonizing}
identified but could not address with their Wilcoxon test, enabling
valid inference for the feature-set comparison.

We report unadjusted $p$-values for the individual DA-vs-baseline
per-source comparisons and additionally assess robustness to a
Bonferroni correction across the four per-source comparisons
($p<0.0125$).

\section{Cross-Source Generalization Results}
\label{sec:results}
 
\subsection{Leave-One-Dataset-Out Performance}
\label{sec:modern}
 
Table~\ref{tab:modern} reports the highest mean VeryHigh recall observed
across the evaluated configurations (rule and std strategies, feature
sets, default and youden\_val threshold rules; 5 seeds each) for each
classifier and held-out source. These maxima are used descriptively to
characterize the strongest observed transfer performance;
 
\begin{table}[t]
\caption{Best observed LODO VeryHigh recall per (classifier, test
source): the highest mean over 5 seeds across the evaluated (strategy,
feature-set, threshold-mode) configurations. These maxima are
descriptive and are not unbiased estimates of a prespecified
deployment configuration.}
\vspace{-0.25cm}
\label{tab:modern}
\centering
\scriptsize
\setlength{\tabcolsep}{4pt}
\begin{tabular}{lcccc}
\toprule
Classifier & Kaggle & Mother Jones & Stanford MSA & GVA \\
\midrule
Random Forest & $0.68 \pm 0.12$ & $0.86 \pm 0.18$ & $0.84 \pm 0.10$ & $0.20 \pm 0.45$ \\
LightGBM      & $0.68 \pm 0.12$ & $0.78 \pm 0.21$ & $0.89 \pm 0.10$ & $0.20 \pm 0.45$ \\
XGBoost       & $0.69 \pm 0.13$ & $0.78 \pm 0.22$ & $0.88 \pm 0.09$ & $0.20 \pm 0.45$ \\
\bottomrule
\end{tabular}
\vspace{-0.25cm}
\end{table}
 
\begin{figure}[!t]
\centering
\includegraphics[width=\columnwidth]{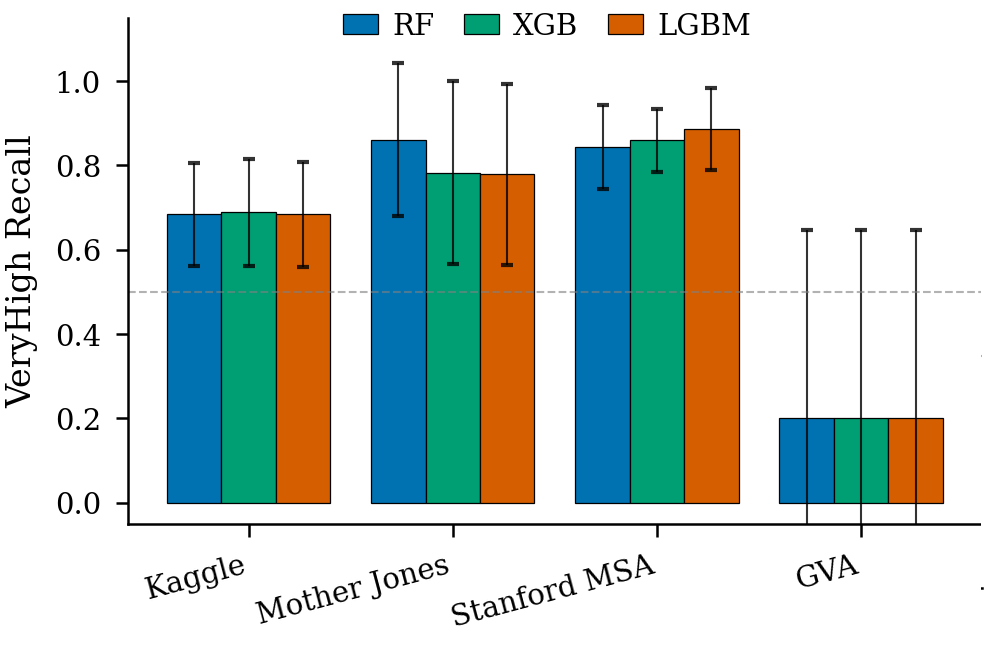}
\vspace{-0.8cm}
\caption{Per-source LODO VeryHigh recall for the best configuration of
each modern classifier. All three collapse on GVA to $0.20 \pm 0.45$
(precision $0.0004$) despite achieving 0.68--0.89 on the curated
sources. Dashed line at 0.5.}
\label{fig:collapse}
\end{figure}
 
All three modern classifiers achieve competitive recall on the three
curated sources (Kaggle 0.68--0.69, Mother Jones 0.78--0.86, Stanford
MSA 0.84--0.89) but collapse on GVA to mean recall of $0.20$ with
standard deviation $\pm 0.45$. VeryHigh precision on GVA is $0.0004$
averaged across all configurations, meaning that when the classifier
predicts VeryHigh on GVA it is almost always wrong. Combined with the
near-zero precision, this indicates that VeryHigh predictions on GVA
are overwhelmingly false positives, with recall varying substantially
across seeds. The large standard deviation reflects seed-level
oscillation between near-zero and near-one recall, indicating unstable
rather than consistently degraded transfer.
 
% \textbf{Illustrative confusion pattern.} To make the extreme precision
% number interpretable: on a representative GVA test fold with std
% labelling and default thresholding, the best-performing modern
% classifier predicts VeryHigh on a large majority of the 6{,}162 GVA
% rows despite only 11 true VeryHigh incidents ($0.18\%$ base rate
% under std labelling), yielding both high recall on the few true
% positives and overwhelming false-positive volume. Averaged across
% all configurations, precision converges to essentially zero because
% most configurations produce this indiscriminate over-prediction.
 
The same failure appears across classifier families and across the tested
strategy, feature-set, and threshold combinations. Two properties of GVA are
particularly important: (i) participant fields are entirely Unknown, so the
contextual and demographic features used by the classifiers are unavailable;
and (ii) GVA's VeryHigh base rate is $10$--$100\times$ lower than curated
sources under both stratification strategies (Table~\ref{tab:baserates}:
$0.03\%$ vs.\ $1.8$--$5.1\%$ under rule; $0.18\%$ vs.\ $11$--$36\%$ under
std). Together, these differences define a severe source shift that coincides
with the observed GVA failure, and motivate the controlled mechanism
analyses in Section~\ref{sec:masking}.
 
Compared with the depth-3 DT best-configuration min-across-source
recall of $0.81$ reported in~\cite{sharma2026harmonizing} on the
3-source pool, none of the modern classifiers reach even the
prior-work floor on the extended 4-source pool. Using modern
classifier families does not resolve the cross-source failure in this
four-source LODO setting.

% \textbf{Choice of metric.} We report VeryHigh recall as the primary
% metric because it aligns with the operational cost model in which
% missing a high-severity event carries substantially higher cost than
% a false alarm, and because it preserves comparability
% with~\cite{sharma2026harmonizing}. Threshold-independent metrics
% (per-class PR-AUC, macro-F1, calibration curves) would provide
% complementary characterization of degradation; the extreme precision
% collapse on GVA ($0.0004$) suggests threshold-tuned metrics would
% similarly show near-random performance, but the specific magnitudes
% are open questions and represent a natural extension.

\textbf{Threshold-independent metrics confirm the GVA collapse.}
To verify that our VeryHigh recall results are not artifacts of
threshold selection, we compute PR-AUC per class using the model's
full probability rankings (rather than any single decision threshold)
on the same LODO folds, reporting the maximum observed value across the
three classifier families (XGBoost, LightGBM, and Random Forest). We additionally report macro-F1 under the default $\arg\max$ threshold as a summary of
overall multi-class performance; note that macro-F1 is
threshold-dependent, while PR-AUC is not.

PR-AUC for the VeryHigh class on GVA is $0.007$. This is $\sim 4\times$
the std-strategy base rate ($0.0018$) and $\sim 20\times$ the
rule-strategy base rate ($0.0003$) (Table~\ref{tab:baserates}), so
the ranking is technically above the random-ranking floor but
effectively unusable: at any recall level that captures even a small
fraction of true positives, precision falls below one percent.
Curated sources achieve substantially higher values: Kaggle $0.402$,
Mother Jones $0.432$, and Stanford MSA $0.155$, between $22\times$
and $62\times$ the GVA value and between $2\times$ and $150\times$
their respective base rates. Macro-F1 follows the same ordering (GVA
$0.25$, curated $0.45$--$0.54$). No configuration among the tested
classifier families and hyperparameter settings achieves PR-AUC
above $0.02$ on GVA. The threshold-independent view therefore
confirms that the GVA collapse reflects genuinely poor class-ranking performance, not a threshold-choice artifact.
 
\subsection{Feature-Set Effect Under Dependence-Aware Inference}
\label{sec:featureset}
 
We compare contextual and full feature sets across all 18
configurations (3 classifiers $\times$ 2 strategies $\times$ 3
threshold modes) using the minimum-across-source recall criterion of
prior work. Because the comparison is paired (ctx vs.\ full at the same
configuration), threshold effects cancel in the ctx$-$full difference,
so spanning all three threshold modes does not affect the validity of
this comparison; the best configuration recalls reported in
Table~\ref{tab:modern}, by contrast, use only the default and
youden\_val rules.
 
The paired ctx-vs-full comparison produces 14 ties, one full-favouring
outcome, and three ctx-favouring outcomes; the mean effect is
$\Delta = +0.018$ (95\% clustered-bootstrap CI $[-0.030, +0.067]$),
and cluster-robust regression gives $\hat{\beta} = +0.018$
(SE $0.025$, $p = 0.464$, 95\% CI $[-0.030, +0.067]$). The high tie
count arises because GVA's near-zero recall floor bounds the
minimum-across-source criterion for most configurations. Under valid
inference on the 4-source pool, we find no statistically detectable effect of the full feature set on minimum-across-source recall; the confidence interval includes both small negative and positive effects, so the present evaluation does not resolve a
reliable advantage for either feature set. This contrasts with the stronger
inference suggested by the configuration-ignoring Wilcoxon analysis
in~\cite{sharma2026harmonizing}.

\subsection{Domain Adaptation}
\label{sec:da}
 
\begin{figure}[!t]
\centering
\includegraphics[width=0.96\columnwidth]{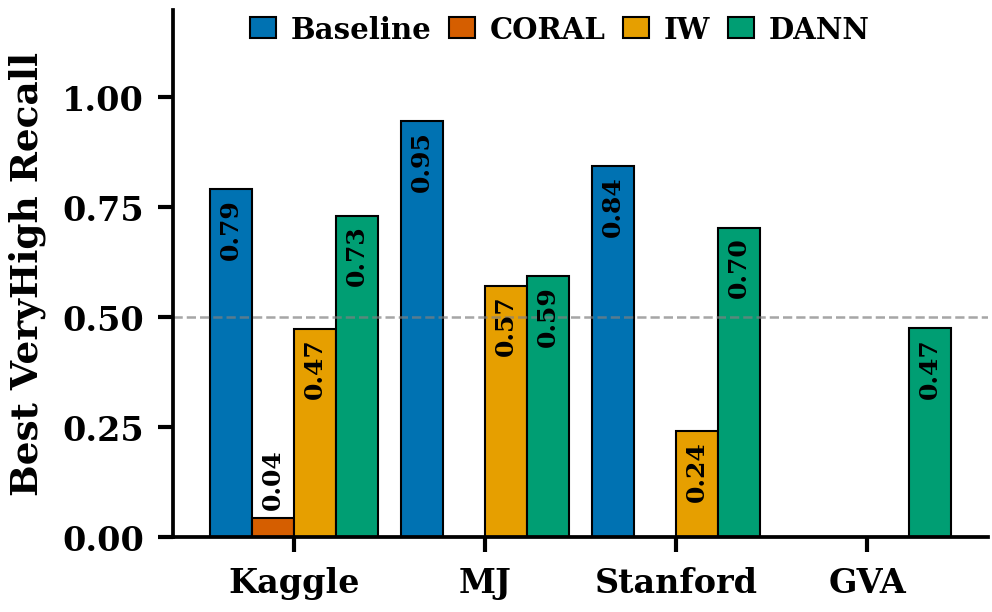}
\vspace{-0.3cm}
\caption{Best observed VeryHigh recall per (test source, method) across the DA
configurations; Table~\ref{tab:da} reports the corresponding grid-averaged values. DANN is the only domain-adaptation method that produces any non-zero recall on
GVA; CORAL and IW hurt every source with informative features.
Baseline is XGBoost without adaptation.}
\label{fig:da}
\end{figure}

Table~\ref{tab:da} reports mean per-source VeryHigh recall for the
baseline XGBoost pipeline and the three domain-adaptation methods;
Fig.~\ref{fig:da} visualizes the per-source comparison.
 
\begin{table}[t]
\caption{Per-source mean VeryHigh recall for baseline XGBoost and three
domain-adaptation methods, averaged across the DA experimental grid
(strategy, features, seeds). Deltas ($\Delta$) are per-source change
relative to baseline. Bold: only positive per-source delta on any
source.}
\vspace{-0.25cm}
\label{tab:da}
\centering
\scriptsize
\setlength{\tabcolsep}{3pt}
\begin{tabular}{lccccc}
\toprule
Method & Kaggle & Mother Jones & Stanford MSA & GVA & Mean \\
\midrule
Baseline XGB & 0.79 & 0.95 & 0.84 & 0.00 & 0.65 \\
CORAL        & $-0.49$ & $-0.68$ & $-0.49$ & $\phantom{+}0.00$ & 0.00 \\
IW           & $-0.33$ & $-0.42$ & $-0.41$ & $\phantom{+}0.00$ & 0.13 \\
\textbf{DANN}
& $\phantom{-}$\textbf{+0.01}
& $-0.22$
& $-0.02$
& $\phantom{-}$\textbf{+0.28}
& \textbf{0.43} \\
\bottomrule
\end{tabular}
\vspace{-0.25cm}
\end{table}
 
\textbf{Baseline definition and cross-table alignment.} The
``Baseline XGB'' row in Table~\ref{tab:da} reports XGBoost's LODO
recall \emph{averaged} over the domain-adaptation experimental grid
(strategy $\in$ \{rule, std\}, features $\in$ \{ctx, full\}, default
thresholding only, 5 seeds), matching the grid over which DA methods
are evaluated. Table~\ref{tab:modern} instead reports the
\emph{best-configuration} XGBoost result, additionally maximizing
over threshold modes (default, youden\_val) --- which is
why, for example, Mother Jones is $0.95$ under the averaged-grid
definition of Table~\ref{tab:da} but $0.78$ under the best-config
definition of Table~\ref{tab:modern}. The DA-comparison baseline uses
the averaged grid because that is the pipeline into which adaptation
methods are inserted; because GVA recall averaged over the DA grid is
exactly zero, the ``help GVA'' question is well-posed.
 
\textbf{Overall pattern.} DANN is the only method that produces any
non-zero VeryHigh recall on GVA (mean $0.28$, best configuration
$0.47$). CORAL and IW both fail to help GVA (zero recall in all
configurations across all five seeds) and hurt every curated source:
CORAL drops recall by $0.49$--$0.68$ across the three curated sources,
and IW drops recall by $0.33$--$0.42$. DANN modestly hurts Mother
Jones ($-0.22$) but is essentially neutral on Kaggle ($+0.01$) and
Stanford MSA ($-0.02$), trading small curated-source performance
losses for a substantial GVA gain.
 
\textbf{Statistical significance under dependence-aware inference.} We apply the
clustered bootstrap and cluster-robust regression of
Section~\ref{sec:inference} to the paired $\Delta$-values, clustering
by (strategy, features). DANN's GVA improvement is statistically
robust: $\Delta = +0.282$ (95\% clustered-bootstrap CI
$[+0.114, +0.469]$; cluster-robust $p = 0.003$). However, DANN's
VeryHigh precision on GVA remains near zero: mean $0.0019$
(best-configuration $0.007$), roughly $5\times$ the baseline precision
of $0.0004$ but still catastrophically low --- when DANN achieves high
recall on GVA it does so by flagging a large majority of the $6{,}162$
GVA rows. DANN therefore improves target-source ranking relative to the baseline, but the resulting operating point remains dominated by false positives;
DANN partially rescues probability ranking but does not move GVA out
of the predict-all-VeryHigh zone.\footnote{A 12-configuration sweep over hidden width, $\lambda_{\max}$, and max epochs yields mean GVA recall $0.19$--$0.43$ (best $0.43$ at hidden=64, $\lambda_{\max}=0.5$, 100 epochs); all 12 are non-zero, while CORAL and IW are zero throughout, and $\lambda_{\max}=0.5$ outperforms $1.0$ ($0.32$ vs.\ $0.25$).}
DANN's curated-source changes are not distinguishable from zero
(Kaggle $\Delta=+0.009$, $p=0.86$; Mother Jones $\Delta=-0.221$,
$p=0.09$; Stanford MSA $\Delta=-0.022$, $p=0.80$), and the aggregate
across-source effect is essentially zero ($\Delta=+0.012$, 95\% CI
$[-0.066, +0.094]$, $p=0.85$): the GVA gain and small curated-source
losses balance --- a recall--precision trade-off in which performance is reallocated from the curated sources toward the hardest target source rather than uniformly improved. In contrast, CORAL produces significant losses on all
curated sources ($p\leq 0.012$; overall $p<0.001$) and zero GVA
recall in every trial; IW shows the same pattern with weaker
significance ($p\approx 0.07$--$0.09$).
 
\textbf{Robustness criterion.} Under the minimum-across-source
robustness criterion of prior work, only DANN produces a
non-zero minimum recall (best configuration 0.47); all other
methods produce a minimum of 0.00, bounded by their GVA failure.
This inverts the pre-adaptation ranking of Table~\ref{tab:modern}, where all
three modern classifiers tied at 0.00 under min-across-source recall.

The complementary median-across-source view makes the trade-off
explicit: baseline XGBoost achieves median $0.82$ (excellent on
curated sources but zero on GVA), DANN median $0.65$ (moderate on
all four), IW median $0.36$, and CORAL median $0.00$. DANN's advantage
under the minimum criterion comes at the cost of median performance,
which is the expected signature of a recall--precision trade-off: the
method reallocates performance from curated sources to the hardest
target rather than uniformly improving.
 
\textbf{Mechanism.} The different behavior of DANN, CORAL, and IW is consistent with how each
method uses target-domain information. CORAL aligns source and target
second-order statistics, while IW reweights source examples using an
estimated target-density ratio. Both approaches therefore depend on target
features that contain useful information about the shift. In GVA, the
demographic and venue attributes are 100\% Unknown, leaving a low-information
feature space. DANN instead learns a domain-invariant representation through
adversarial training. This difference in how target-domain information
is used is consistent with DANN retaining some GVA signal when the target features are systematically incomplete.

\section{Mechanism Analysis}
\label{sec:mechanism}

The preceding results establish that cross-source generalization
fails on GVA and that domain-adversarial training partially recovers
it. This section examines the possible mechanisms through two complementary
experiments: feature masking tests whether incomplete feature
availability can reproduce the failure, and prior-shift recalibration
tests whether label-prevalence mismatch alone is sufficient to explain it.

\subsection{Feature-Masking Ablation}
\label{sec:masking}

To test whether feature completeness (rather than temporal
drift or joint-distribution differences) contributes to the
cross-source failure, we run a controlled ablation. We mask the five
features that GVA lacks (\texttt{gender}, \texttt{race},
\texttt{mental\_health}, \texttt{incident\_area}, \texttt{age}) on
\emph{all} sources by setting them to \texttt{Unknown} or NaN, then
re-run the baseline LODO pipeline (XGBoost, std strategy, full
features, 5 seeds). This holds temporal, base-rate, and marginal-distribution
properties constant across the two conditions --- only feature
availability changes. We use the same fixed baseline configuration in
both conditions to isolate the effect of feature availability.

Table~\ref{tab:masking} provides controlled evidence for the
feature-completeness hypothesis: masking the five features unavailable
in GVA collapses VeryHigh recall on the curated sources to zero.

\begin{table}[t]
\centering
\caption{Feature-masking ablation under the best baseline LODO
configuration (XGBoost, std strategy, full features); mean VeryHigh
recall $\pm$ std over 5 seeds.}
\label{tab:masking}
\begin{tabular}{lcc}
\toprule
Source & Baseline recall & Masked recall \\
\midrule
Kaggle & $0.544 \pm 0.035$ & $0.000 \pm 0.000$ \\
Mother Jones & $0.643 \pm 0.013$ & $0.000 \pm 0.000$ \\
Stanford MSA & $0.795 \pm 0.024$ & $0.000 \pm 0.000$ \\
GVA & $0.000 \pm 0.000$ & $0.745 \pm 0.297$~$^{\dagger}$ \\
\bottomrule
\end{tabular}
\parbox{\linewidth}{\footnotesize $^{\dagger}$Degenerate predict-all-VeryHigh regime; precision $0.002$.}
\end{table}

The collapse is robust: it appears under both \texttt{rule} and
\texttt{std} strategies and both feature sets. GVA recall increases to
$0.745$ under masking, but this is a degenerate predict-all-VeryHigh
regime with precision of only $0.002$; the apparent recall gain does
not represent recovered discriminative information, but rather a
collapse toward a trivial majority-action strategy once the remaining
features are removed. Because the masked representation provides little
usable per-instance information, and temporal drift and
joint-distribution shift are held constant by design, this provides
controlled evidence for the failure mode discussed in
Section~\ref{sec:discuss}.

% \textbf{Lowest-common-denominator baseline.} A complementary question
% is whether \emph{any} predictive signal exists using only features GVA
% reliably provides. Under our leakage-avoiding feature definition (which
% excludes \texttt{fatalities}, \texttt{injured}, \texttt{total\_victims},
% and \texttt{year}), the intersection of GVA-available and paper-eligible
% features is essentially empty: \texttt{gender}, \texttt{race},
% \texttt{age}, \texttt{mental\_health}, and \texttt{incident\_area} are
% all Unknown on GVA, and \texttt{multiple\_shooters} is inferred from a
% derived suspect count. This structural constraint --- rather than any
% modeling choice --- is the fundamental reason cross-source transfer to
% GVA fails: the target domain provides no informative features that
% carry non-outcome signal.

\subsection{Prior-Shift Recalibration}
\label{sec:labelshift}

Because base-rate mismatch is one of the two mechanisms we hypothesize
for the GVA collapse (Table~\ref{tab:baserates}), a natural
question is whether label-shift correction alone can recover GVA
performance. We evaluate prior-shift recalibration
(Saerens et al., 2002; also the closed-form component
of BBSE~\cite{lipton2018bbse}) as the standard label-shift-only
baseline: given trained classifier posteriors
$P_{\text{source}}(y\!\mid\!x)$ and known target class prior
$P_{\text{target}}(y)$, compute
$P_{\text{target}}(y\!\mid\!x) \propto P_{\text{source}}(y\!\mid\!x)
\cdot P_{\text{target}}(y) / P_{\text{source}}(y)$, then take
$\arg\max$. We use the observed test-source class distribution as
$P_{\text{target}}(y)$, giving an oracle upper bound on what any
label-shift-only method could achieve.

Table~\ref{tab:labelshift} reports mean VeryHigh recall and precision
per source, averaged over five seeds and the same configuration grid
used for the DA experiments. Two findings emerge.

\begin{table}[t]
\centering
\caption{VeryHigh-class performance before and after oracle
prior-shift recalibration; mean over 5 seeds and the same
configuration grid used for the DA experiments.}
\label{tab:labelshift}
\small
\begin{tabular}{lcccc}
\toprule
& \multicolumn{2}{c}{Baseline XGB} & \multicolumn{2}{c}{Prior-shift} \\
\cmidrule(lr){2-3} \cmidrule(lr){4-5}
Source & Recall & Precision & Recall & Precision \\
\midrule
Kaggle & $0.342$ & $0.251$ & $0.253$ & $\mathbf{0.373}$ \\
Mother Jones & $0.557$ & $0.269$ & $0.171$ & $0.278$ \\
Stanford MSA & $0.160$ & $0.172$ & $0.040$ & $0.079$ \\
GVA & $\mathbf{0.000}$ & $\mathbf{0.000}$ & $\mathbf{0.000}$ & $\mathbf{0.000}$ \\
\bottomrule
\end{tabular}
\end{table}

First, \textbf{prior-shift recalibration does not recover any VeryHigh
signal on GVA}: both recall and precision remain at exactly zero across
all five seeds and all configurations. This is the paper's cleanest
empirical support for the feature-completeness mechanism as
\emph{primary}: correcting for the known base-rate mismatch --- with
oracle access to the target prior --- does not help, because the
underlying classifier posteriors on GVA carry no discriminative
information to reweight. Label-side correction cannot rescue feature-side collapse.
Because we use the oracle target prior, non-oracle estimators such as
BBSE~\cite{lipton2018bbse} can only match or underperform this result;
BBSE's confusion-matrix inversion is additionally ill-conditioned
under the GVA collapse (recall $=0$ across classes).

Second, on sources where features carry signal, recalibration behaves
as expected: it shifts the precision--recall trade-off toward higher
precision at the cost of recall (Kaggle precision $0.25\!\to\!0.37$,
recall $0.34\!\to\!0.25$), confirming correct implementation.

\section{Fairness and Responsible Evaluation}
\label{sec:fairness}
 
Fairness has been extensively studied in the Big Data community from
both generative and algorithmic angles~\cite{xu2018fairgan,barocas_selbst};
our contribution is diagnostic rather than corrective, identifying
the mechanism by which cross-source model behaviour disproportionately
affects mental-health-flagged incidents. We audit per-group VeryHigh
recall and precision under the best modern LODO configuration
(XGBoost + std + ctx + youden\_val) for the three available demographic
attributes: gender, race, and mental\_health.
 
\subsection{Coverage Constraints}
 
Gender-based auditing is uninformative across all four sources due to
$\sim 95$\% male-perpetrator prevalence. Racial composition is
majority-Unknown in Kaggle (65\%) and Stanford MSA (73\%), leaving
only Mother Jones with substantial racial variation but the smallest
overall sample ($n=157$). GVA lacks all participant demographic
information (100\% Unknown for both attributes); combined with the
classifier's collapse on GVA (Section~\ref{sec:modern}), this
precludes meaningful group-conditioned analysis on the largest source.
 
\subsection{Mental-Health-Associated Prediction Disparities}
 
\begin{figure}[!t]
\centering
\includegraphics[width=0.96\columnwidth]{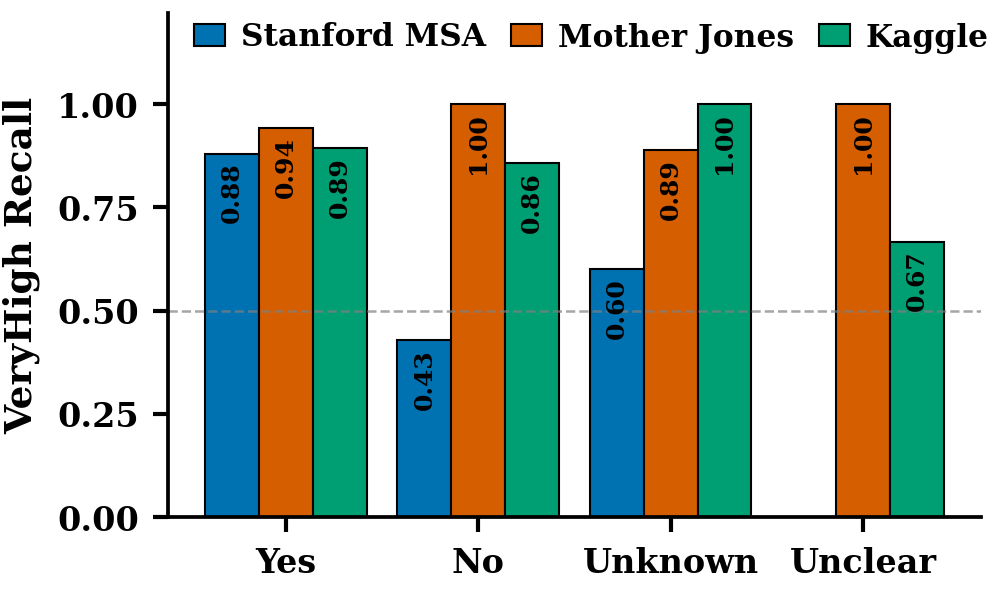}
\vspace{-0.3cm}
\caption{Per-group VeryHigh \emph{recall} by \texttt{mental\_health}
label across the three curated sources. Incidents flagged
\texttt{Yes} on Stanford MSA and Kaggle receive substantially higher
recall than \texttt{No} (indicating over-prediction); Mother Jones
shows near-uniform high recall across groups. Stanford MSA lacks an
\texttt{Unclear} category (only Yes/No/Unknown are recorded in its
schema). Precision values are reported in Table~\ref{tab:fairness_ci}.
GVA is omitted (all participant fields are Unknown, precluding
group-wise audit).}
\label{fig:fairness}
\end{figure}

On Stanford MSA and Kaggle, the classifier assigns systematically
higher VeryHigh recall to incidents labelled
\texttt{mental\_health=Yes} than to \texttt{mental\_health=No}
(Stanford: $0.88$ vs $0.43$, a 45-point gap; Kaggle: $0.89$ vs $0.86$,
a 3-point gap). Table~\ref{tab:fairness} decomposes the Stanford
result, and Fig.~\ref{fig:fairness} visualizes both sources with
recall and precision side by side. Elevated recall on the Yes group
may appear favourable in isolation, but precision on the same group
drops to $0.34$ (Stanford) and $0.31$ (Kaggle): approximately two
thirds of VeryHigh predictions on mental-health-flagged incidents
are false positives.
 
\begin{table}[t]
\caption{Per-group VeryHigh recall and precision on Stanford MSA for
the best modern LODO configuration (XGBoost + std + ctx +
youden\_val). ``Yes'' incidents receive substantially higher
predicted-risk rates, but with precision collapse.}
\vspace{-0.25cm}
\label{tab:fairness}
\centering
\scriptsize
\begin{tabular}{lccccc}
\toprule
Group & $n$ & $n_{pos}$ & Recall & Precision & TP / FP / FN \\
\midrule
Yes     & 96 & 25 & 0.88 & 0.34 & 22 / 43 / 3 \\
Unknown & 145 & 5 & 0.60 & 0.08 & 3 / 34 / 2 \\
No      & 93 & 7 & 0.43 & 0.08 & 3 / 37 / 4 \\
\bottomrule
\end{tabular}
\vspace{-0.25cm}
\end{table}
 
Effectively, the classifier learns \texttt{mental\_health=Yes} as a
proxy for elevated risk, over-predicting VeryHigh for this group at
three to four times the observed base rate. This pattern is consistent
with disparate prediction behavior associated with the
\texttt{mental\_health} attribute, of the type warned about
in~\cite{barocas_selbst}: operational deployment would predominantly
generate false-positive high-risk flags for incidents already
carrying mental-health labels, a population already
disproportionately affected by surveillance and stigmatizing
downstream interventions.
 
Mother Jones's mental\_health=Yes rate is $51\%$ vs.\ Kaggle ($33\%$)
and Stanford MSA ($29\%$), reflecting editorial focus rather than
clinical prevalence; the training signal partly encodes which
incidents journalists cover. The flag is media-derived in all four
sources, further weakening causal interpretation.

\textbf{Uncertainty and reverse causality.} Wilson $95\%$ confidence
intervals on per-group VeryHigh recall and precision (Table~\ref{tab:fairness_ci})
show the Yes-vs-No gap on Stanford MSA is not attributable to sampling
variability: Yes recall $0.88$ $[0.69, 0.97]$ vs.\ No recall $0.43$
$[0.10, 0.82]$; Yes precision $0.34$ $[0.23, 0.47]$ vs.\ No precision
$0.07$ $[0.02, 0.20]$. The intervals overlap on recall only at their
extremes, and precision intervals are strictly disjoint.

A reverse-causality concern qualifies the disparate-impact interpretation:
media outlets may selectively assign a \texttt{mental\_health=Yes} label
to incidents that are already more severe (retrospective attribution
following high-fatality attacks), which would inflate the observed
correlation between the flag and VeryHigh outcomes independent of
classifier behavior. Our observational data cannot distinguish reverse
causality from a genuine classifier-side disparate impact; disentangling
these would require either a randomized labeling audit (assigning
\texttt{mental\_health} status blind to outcome) or access to the
timing of label attribution relative to outcome reporting, neither of
which is available in the source databases. The disparate-impact
finding should therefore be read as identifying a joint
labeling-plus-classifier failure mode, not as an attribution to the
classifier alone.

\begin{table}[t]
\centering
\caption{Wilson $95\%$ CIs on per-group VeryHigh recall and precision
(\texttt{mental\_health} attribute).}
\label{tab:fairness_ci}
\small
\setlength{\tabcolsep}{3pt}
\begin{tabular}{llcc}
\toprule
Source & Group & Recall (95\% CI) & Precision (95\% CI) \\
\midrule
Stanford MSA & Yes     & $0.88$ $[0.69, 0.97]$ & $0.34$ $[0.23, 0.47]$ \\
Stanford MSA & No      & $0.43$ $[0.10, 0.82]$ & $0.07$ $[0.02, 0.20]$ \\
Stanford MSA & Unknown & $0.60$ $[0.15, 0.95]$ & $0.08$ $[0.02, 0.22]$ \\
Kaggle       & Yes     & $0.89$ $[0.72, 0.98]$ & $0.31$ $[0.21, 0.42]$ \\
Kaggle       & No      & $0.86$ $[0.42, 1.00]$ & $0.11$ $[0.04, 0.22]$ \\
Kaggle       & Unknown & $1.00$ $[0.48, 1.00]$ & $0.17$ $[0.06, 0.35]$ \\
\bottomrule
\end{tabular}
\end{table}
 
% \subsection{Sample-Size Caveats}
 
% Racial subgroup sizes in Kaggle ($n$ between 6 and 204) and Stanford
% MSA ($n$ between 0 and 145) preclude statistically stable
% single-attribute audits. The observed race gaps should be interpreted
% as descriptive rather than confirmatory.
 
\subsection{Fairness Implications of Cross-Source Failure}
 
The GVA collapse (Section~\ref{sec:modern}) constitutes a fairness
concern beyond individual-attribute gaps: a model that fails to make
correct predictions on the majority of its target population cannot be
meaningfully audited for group-level disparities on that population.
Models that do not work cannot be audited for whether they work
equitably.
 
% ============================================================
\section{Discussion}
\label{sec:discuss}
% ============================================================
 
\subsection{Why Cross-Source Generalization Fails}
 
Our results point to a clear failure pattern. Modern classifiers
trained on the three curated sources rely on contextual (venue) and
demographic (mental\_health, race) features for VeryHigh classification;
these features do not transfer to GVA because its participant fields
are systematically Unknown. Meanwhile, GVA's VeryHigh base rate is
substantially lower than the curated sources under both stratification
strategies (Table~\ref{tab:baserates}: rule $0.03\%$ vs.\ $1.8$--$5.1\%$;
std $0.18\%$ vs.\ $11$--$36\%$), a base-rate mismatch that no threshold
rule tested here can correct.
 
This mechanism has implications for how the mass-shooting ML
literature should be read. Prior single-source models trained on a
curated dataset are likely to appear substantially more performant
than they truly are on the general mass-shooting population, because
the curated sources have systematically different casualty
distributions, participant coverage, and inclusion criteria than the
largest available public database.

\textbf{On isolating the mechanism.} Feature completeness and
base-rate shift co-vary with temporal and distributional differences
across sources. Three independent lines of evidence support feature
completeness as a contributor: (i)~GVA's joint MMD$^2$ against
curated sources ($\sim 0.14$) is smaller than the
Kaggle$\leftrightarrow$Mother Jones MMD$^2$ ($0.243$), yet GVA transfer
fails while Kaggle$\leftrightarrow$Mother Jones transfer succeeds
(Section~\ref{sec:shift}); (ii)~only DANN's feature-invariant
approach recovers any GVA signal (Section~\ref{sec:da}); and (iii)~masking
the five features GVA lacks collapses curated-source recall from
$0.54$--$0.80$ to zero (Section~\ref{sec:masking}). Complemented by the
null oracle prior-shift result (Section~\ref{sec:labelshift}), these
indicate that feature completeness is a contributor and that
base-rate mismatch alone does not explain the collapse.
 
\subsection{Why Simple DA Fails and DANN Succeeds}
 
The differential success of DANN illuminates the limits of
alignment-based domain adaptation. CORAL and IW both assume the target
has informative features to align against or reweight toward. GVA's
Unknown-dominated feature matrix violates this assumption: the
second-order statistics CORAL aligns are almost entirely determined by
the Unknown categorical values, and the density ratios IW estimates
carry essentially no signal because the marginal distributions of
source and target diverge in a region where the classifier has learned
no discriminative representation.
 
Gradient-reversal training does not require target features to be
individually informative; it requires only that a domain-invariant
subspace exists in which the label signal survives. This is consistent
with DANN retaining non-zero GVA signal while CORAL and IW, both of which
require informative target features to align against or reweight
toward, do not.

\textbf{Feature parity.} All DA methods (DANN, CORAL, IW) and baseline
classifiers use identical feature sets (\texttt{ctx} or \texttt{full}
as defined in Section~\ref{sec:setup}) with outcome variables
(\texttt{fatalities}, \texttt{injured}, \texttt{total\_victims},
\texttt{year}) strictly excluded from all pipelines. DANN has no
informational advantage over the baseline classifiers; the only
difference is the adversarial domain-invariance objective.
 
This is a positive but modest result. DANN's best configuration
achieves GVA recall of $0.47$, meaning more than half of true VeryHigh
incidents in GVA are still missed. The result should be read as
``adversarial DA is the only tested method that can produce any signal
on GVA transfer,'' not as ``cross-source generalization is solved.''
 
\subsection{Limitations}
\label{sec:limitations}
Our extended benchmark is bounded by six limitations. First, harmonization requires many source-specific mapping decisions (particularly for \texttt{incident\_area} and \texttt{mental\_health}); reasonable choices differ across teams. Second, GVA's Unknown-heavy feature landscape reflects a real information constraint, but our GVA results are contingent on the specific imputation strategy. Third, all four sources inherit media-coverage biases that no ML method can correct; the benchmark measures generalization across biased sources, not fidelity to the underlying phenomenon. Fourth, DANN uses a two-layer MLP while CORAL and IW use XGBoost, partially conflating adversarial adaptation with base-learner architecture; the masking ablation (Section~\ref{sec:masking}) provides indirect support that the mechanism is not purely architectural. Fifth, quartile-based labelling degenerates on the extended pool (Section~\ref{sec:strat}), so prior-work comparison uses only rule and std strategies. Sixth, we evaluate one label-shift correction but not related methods (full BBSE~\cite{lipton2018bbse}, CDAN, calibration diagnostics, ordinal-severity formulations, or richer missingness treatments); our oracle prior-shift result indicates pure label-side correction is insufficient on GVA, and feature-side methods appear necessary for meaningful progress.
 
% \subsection{Ethics and Non-Deployment}
 
% This pipeline is a research artifact, not a deployable system.
% Operational use would inherit the coverage biases we document, would
% amplify the mental-health disparate impact quantified in
% Section~\ref{sec:fairness}, and would provide false-positive risk
% assessments at high rates on populations already disproportionately
% surveilled. The harmonized schema and all code are released under a
% non-commercial research license that explicitly prohibits operational
% deployment.
 
% We follow the ethical framing of~\cite{barocas_selbst}: the
% mental\_health, race, and gender features in our schema are
% media-attributed proxies, not clinical or self-identified attributes,
% and should not be interpreted causally in any downstream use of this
% dataset.
 
% ============================================================
\section{Conclusion}
% ============================================================
 
MASH-Bench extends prior cross-source mass-shooting risk classification
work~\cite{sharma2026harmonizing} with a fourth source (GVA), modern
classifier baselines, three DA methods, dependence-aware inference
under configuration dependence, a feature-masking mechanism ablation,
oracle label-shift correction, and a per-group fairness audit. Cross-source generalization degrades sharply on GVA under conditions
where scaling classifier capacity alone does not help: feature-completeness
gaps and $10$--$100\times$ lower VeryHigh base rate
(Table~\ref{tab:baserates}) together produce a failure pattern that
neither modern classifiers, alignment-based DA, nor oracle
label-shift correction reduces in our experiments. DANN is uniquely non-zero
on GVA ($\Delta=+0.28$, $p=0.003$) but partial and reallocative rather
than uniformly improving. Per-group auditing reveals that mental-health-flagged
incidents receive systematically over-predicted high-risk assignments
with $\sim 34\%$ precision --- a disparate prediction pattern that
operational deployment would amplify. MASH-Bench is a research artifact,
not a deployable system: operational use would amplify the documented
disparate prediction behavior on media-flagged mental-health cases.
We release the harmonized schema and code under a non-commercial research license.

\bibliographystyle{IEEEtran}
\bibliography{references}

\end{document}